# Action Recognition: From Static Datasets to Moving Robots

Fahimeh Rezazadegan, Sareh Shirazi, Ben Upcroft and Michael Milford*

*Abstract*— Deep learning models have achieved state-of-the-art performance in recognizing human activities, but often rely on utilizing background cues present in typical computer vision datasets that predominantly have a stationary camera. If these models are to be employed by autonomous robots in real world environments, they must be adapted to perform independently of background cues and camera motion effects. To address these challenges, we propose a new method that firstly generates generic action region proposals with good potential to locate one human action in unconstrained videos regardless of camera motion and then uses action proposals to extract and classify effective shape and motion features by a ConvNet framework. In a range of experiments, we demonstrate that by actively proposing action regions during both training and testing, state-of-the-art or better performance is achieved on benchmarks. We show the outperformance of our approach compared to the state-of-the-art in two new datasets; one emphasizes on irrelevant background, the other highlights the camera motion. We also validate our action recognition method in an abnormal behavior detection scenario to improve workplace safety. The results verify a higher success rate for our method due to the ability of our system to recognize human actions regardless of environment and camera motion.

## I. INTRODUCTION

Recognizing and understanding human activity is essential for a wide variety of applications from surveillance purposes [1] and anomaly detection [2] to having safe and collaborative interaction between humans and robots in shared workspaces. More explicitly, for robots and humans to be cooperative partners that can assist human intuitively, it is crucial that robot recognizes the actions of human. With such abilities, a robot can identify the next required task to assist a human at the appropriate time as well as reducing the likelihood of interfering with the human activity [3].

Over the last decade, significant progress has been made in the action recognition field using conventional RGB images, optical flow information and the fusion of both [4]. Transitioning these computer vision techniques from benchmark dataset to real world robots is challenging. Real world imagery is far more diverse, unbiased and challenging than computer vision datasets, meaning these techniques tend to perform far worse when applied blindly to a robot vision system [5].

Transitioning from computer vision approaches to robotics applications involves two main challenges. Firstly, the computer vision approaches rely on background cues due to the fact that traditional datasets tend to have contextually-informative backgrounds. Secondly, having datasets that mainly use stationary cameras would make the methods vulnerable to disturbing effects of camera motion.

* The authors are with the Australian Centre for Robotic Vision (ACRV), School of Electrical Engineering and Computer science, Queensland University of Technology, Brisbane, Australia. email: {firstname.lastname@qut.edu.au}.

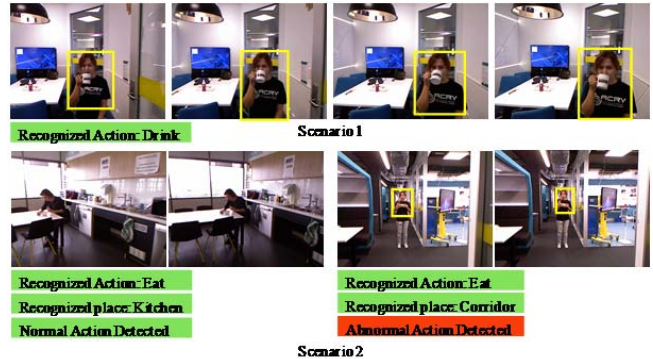

Figure 1. Performance of our action recognition approach in two scenarios. Scenario1 involves action recognition by a moving robot with unbiased background. Scenario 2 comprises abnormal behavior detection in an office environment.

This would negatively impact the performance in robotics applications where it is critical to have mobile platforms.

Motivated by the benefits of using object proposals in object recognition, it is demonstrated that generation of action region proposals is of great importance, because we can focus on the motion salient regions rather than the full video frames [7]. This leads to a big reduction in computational cost and an improvement in performance due to elimination of the background cues [6],[7]. However, to the best of our knowledge, no work has addressed two aforementioned challenges simultaneously.

In this paper, we develop an action recognition system, that recognizes human actions regardless of the platform, background context and camera motion by jointly detecting and recognizing actions based on a new action region proposal method. To this end, we firstly correct the temporal cues by removing the effect of camera motion and then exploit the human motion boundaries to select a reliable action region proposal that are fed to the Convolutional Neural Networks (ConvNet). Through a wide range of experiments, we test our algorithm on 1) benchmark dataset [8], 2) a new datasets containing non-informative background, 3) a new dataset recorded by a mobile robot. We also validate our system in an abnormal human behaviour detection scenario to improve the workplace safety, which is applicable to other fields such as improving elderly care and reducing driving risk [9]. The approach in this experiment detects the abnormal actions in the work environment by jointly categorizing the scene and recognizing actions (Figure 1). Our paper provides the following contributions:

- We develop a new framework for jointly detecting and recognizing human activities using novel action region proposals. This enables categorization which is robust against both camera motion and irrelevant background contexts, and is therefore suitable for robots operating in the real world.

- We introduce two new unbiased datasets (without background bias); one achieved through careful composition of camera footage, the other through acquisition by a mobile robot.

- We conduct a comprehensive suite of experiments evaluating the performance of our proposed technique on two benchmark datasets and the new unbiased background datasets.

- We evaluate the performance of the proposed approach against existing state-of-the-art methods on our dataset recorded by a mobile robot to recognize human actions in work environment on our university's campus.

- Based on our action recognition system, we introduce an abnormal behavior detection scenario, in which the robot is able to detect abnormal behaviors.

The rest of paper is organized as follows. In Section II, we review related work on action recognition in robotics and computer vision fields. We then present an overview of the approach and describe our network architectures in Section III. Section IV details experiment setup and experimental results followed by conclusion in Section V.

## II. RELATED WORK

In robotics, action recognition plays a critical role for fluent human-robot interactions. There has been a number of studies on human action recognition [1], [10], and prediction [2]. Both hand crafted local feature representations and deep learned feature descriptors have been employed in these approaches, with both categories demonstrating excellent results in recognition of human actions. Hand-crafted local features such as Space Time Interest Points [11], Cuboids [12], Dense Trajectories [13], with rich descriptors of HOG, HOF, MBH have shown to be successful on a number of challenging datasets [8], [14].

Although motion is an informative cue for action recognition, irrelevant motions in the background or the camera motion can be misleading. This is inevitable when dealing with realistic robotic applications in uncontrolled settings. Therefore, separating human action motion from camera motion remains a challenging problem. A few number of works tried to address this isse. Ikizler-Cinbis et al. utilized video stabilization by motion compensation for removing camera motion [15]. Wu et al. addressed the camera motion effects by decomposing Lagrangian particle trajectories into camera-induced and object-induced components for videos [16]. Wang et al. proposed a descriptor based on motion boundary histograms (MBH) which removes constant motions and therefore reduces the influence of camera motion [13]. What makes our method different from [13], is that we first reduce the smooth camera motion effects and get rid of background clutter by creating action region proposals based on a motion boundary detector. The selected regions would be used both in training and classification. However, the approach in [13] employs MBH on full images as motion descriptor for trajectories.

Among traditional methods, there are very few works that have tried to separate the background clutter from images. Chakraborty et al. presented an approach based on selective Spatio-Temporal Interest Points (STIPs) which are detected by suppressing background SIPs and imposing local and temporal constraints, resulting in more robust STIPs for actors and less unwanted background STIPs [17].

Zhang et al. addressed the activity recognition problem for multi-individuals based on local spatio-temporal features in which extracting irrelevant features from dynamic background clutter has been avoided using depth information [10]. Our work is different from them in terms of jointly eliminating background clutter and camera motion using optical flow and motion boundary detection concept.

Deep learning models are a class of machine learning algorithms that learn a hierarchy of features by building high-level features from low-level ones. After impressive results of ConvNets on image classification tasks [18], researchers have also focused on using ConvNet models for action recognition. Several outstanding techniques are introduced that have had a significant impact on this field, such as 3D CNNs [19], RNN [20], CNNs [21] and Two-Stream ConvNet [22].

The majority of recent research has employed motion information to improve the results. Simonyan and Zisserman proposed a two stream ConvNet [22], which has formed the baseline of more recent studies [20]. In [22], spatial and temporal networks are trained individually and then fused. Additionally, two different types of stacking techniques are implemented for the temporal network, optical flow stacking and trajectory stacking. These techniques stack the horizontal (x) and vertical (y) flow channels ($d_t^{x,y}$) of L consecutive frames to form a total of 2L input channels and obtained the best result for L=10 or 20-channel optical flow images. Recently, building on top of traditional Recurrent Neural Networks (RNNs), Donahue et al. proposed a long-term recurrent convolutional model that is applicable to visual time-series modeling [20].

However, deep models ignore the effect of background dependency and moving camera in their training process and evaluations. In this work, our system is able to cope with the background clutter as well as camera motion using several motion cues to eliminate the regions that do not contain the human action.

## III. OVERVIEW OF THE SYSTEM

Our human action recognition approach consists of two main stages:

1) Selecting the action region proposals (motion salient regions) independent of camera motion and background information.

2) Training ConvNets on action region proposals both in spatial and optical flow images, rather than full images. In the training process, we used 3 different ConvNet architectures: two stream ConvNet [22] followed by an SVM classifier to fuse the spatial and temporal features, a 3-D ConvNet that classifies a sequence of video frames as a video clip [23] and a very deep convolutional neural network [24] which is employed under the same two-stream framework.

The summary of approach is visualized in Figure 2. We describe each part in the following, before presenting experiments and evaluations in the next section.

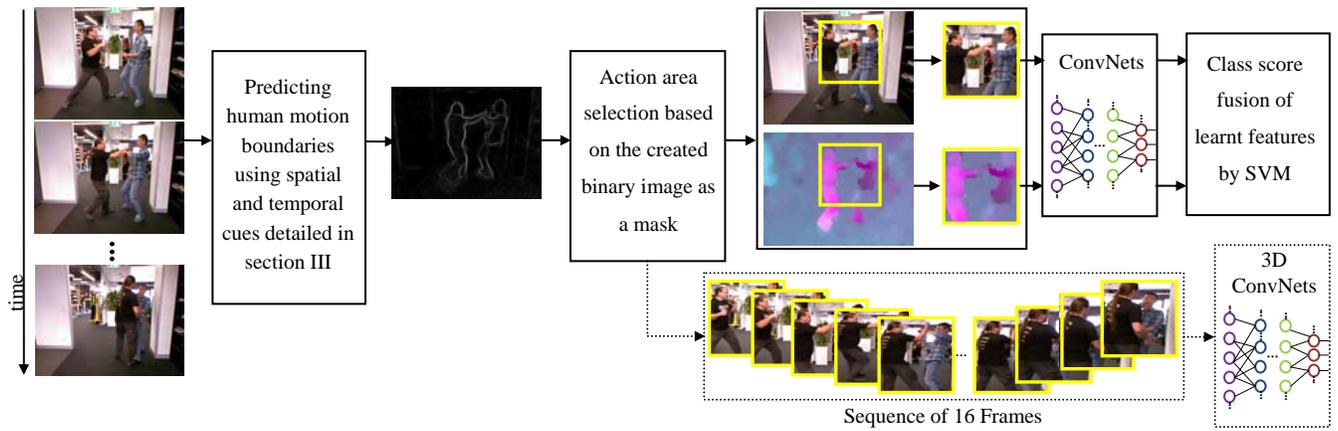

Figure 2. Overview of our approach for unbiased human action recognition on samples of the Guiabot robot dataset. The robot is moving from left to right, while approaching to people. The method is tested using two different ConvNet architecture, denoted by solid and dotted blocks.

## A. Selecting Action Region Proposals

Choosing the action region would eliminate irrelevant regions, which reduces the number of regions being processed, and subsequently faster computation time. However, we face some challenges to have a precise action region proposal. The main challenge of choosing action region proposals compared to object proposals, is that we require both appearance and motion cues to be able to select the motion salient area. Differentiating human actions from the background or other dynamic motions is the first challenge due to the diversity of human actions. The second challenge would be caused by a moving camera. In many computer vision systems, data are only recorded by stationary cameras, which is unlikely the case in robotics applications. Therefore, it is essential to be able to handle camera motion.

In order to handle the mentioned challenges, we leverage the concept of motion boundaries to pick the interested area that only contains human activity. We firstly generate a mask by computing the motion boundaries using an algorithm that is built upon the presented work in [25]. Then we extract the action region proposals from video frames using the previously generated mask followed by an object proposal method [25, 26].

To generate the motion boundaries, we use a combination of different spatial and temporal cues to shape a robust feature representation. The spatial information is three RGB channels, the norm of the gradient and the oriented gradient maps in four directions at coarse and fine scales.

We use multiple temporal cues to identify motion boundaries and generate our cropping mask. The first cue is the horizontal and vertical optical flow signals for both forward and backward process, computed by the state-of-the-art algorithm, classic+NL, proposed in [28] due to the sharpness of the flow boundaries which results in the best optical flow performance. The second one would be an unoriented gradient map computed as the magnitude of horizontal and vertical optical flow gradient maps. The third temporal cue is oriented gradient maps in four directions at a coarse scale computed as the average of gradient maps components, weighted by their magnitudes. The next cue would be image warping errors which can be critical to prevent some optical flow estimation faults. We can compute image warping errors $E_D$ using (1) which is defined at a pixel p as

$$E_D(p) = \|D_t(p) - D_{t+1}(p + W_{t;t+1}(p))\|_2 \quad (1)$$

Where $W_{t;t+1}$ is optical flow between frame $t$ and $t+1$ and $D$ is a pixel-wise histogram of oriented gradients in eight orientations, which are all individually normalized to unit norm. The last one is motion boundaries histogram (MBH) that represents the gradient of the optical flow and can remove locally constant camera motion while keeping information about changes in the flow field. We compute spatial derivatives for both horizontal and vertical optical flow and orientation information is quantized into histograms, while we use the magnitude for weighting.

Given this feature, we predict the binary boundary mask using structured random forests such that the predicted masks are averaged across all trees and all overlapping patches to yield the final soft-response boundary map [25]. Then, we employ it as a mask for video frames such that the area of motion is highlighted. Inspired by object detection approaches [26], [27], we select the desired region by applying an object detection method [26] on the resulted mask with highlighted motion areas.

In the following sections, we explain the procedure for the training and classification, which are done using three different ConvNet architectures.

## B. Training Process and Classification

Recently proposed methods train the network by center cropping or randomly cropping the full image [22], [20], [10]. As a result, these approaches might fail in real robotic scenarios due to confusion caused by unbiased background and a moving camera. Conversely, our approach addresses those challenges by automatically identifying the image region where the action is likely to occur and then passes the action region as the input to the network. This process ensures that the most pertinent information to action is utilized. Therefore, we extract motion and appearance features of the motion salient region even if the actor's spatial location changes throughout the image.

### 1) Training a 16-Layer ConvNet

We train our spatial and temporal networks on action region proposals obtained from Section A in spatial and temporal domains, respectively. Then we concatenate learnt features from both spatial and temporal streams and pass it to a SVM classifier to have a final classification. Our spatial and temporal networks contain three convolutional layers, three pooling layers and two fully connected layers that is built on top of the VGG-16 Layers architecture [29] implemented in Caffe [30].

### 2) Training a 3D ConvNet

We also actively train on a sequence of our proposed RGB images using C3D architecture, which is particularly a good feature learning machine for action recognition [23]. We use 5 convolution layers, followed by 5 pooling layers, 2 fully-connected layers and a softmax loss layer for predicting action labels. The number of filters for 5 convolution layers are 64, 128, 256, 256, 256, respectively. We input 16 frames as a video clip for each video either in benchmark or our introduced datasets with the kernel size of 3 as the temporal depth due to verified experimental results in [23]. As a result, the input dimension for training on our action proposals equals to $3\times16\times112\times112$. Since the 3D architecture involves exploiting both spatial and temporal cues during the training process, no temporal network is required.

### 3) Training a 152-Layer ConvNet

Another inspiring architecture to apply our method is ResNet which is introduced recently [24]. To the best of our knowledge, this architecture has not been used for action recognition, while we have found it so effective in this task. Residual network can overcome the degradation problem through direct identity mappings between layers as skip connections, which allow the network to pass on features smoothly from earlier layers to later layers. We feed our cropped spatial and optical flow images from Section A, which are resized to $224\times224$, to our network containing 152 layers including convolutional layers and skip connections ending with a global average pooling layer and a fully-connected layer with softmax.

## IV. EXPERIMENTAL SETUP

In this section we briefly explain our validation setup on benchmarks and three other experimental setups.

### A. Validation on Benchmarks

To have a thorough investigation of our method, we applied our method on two benchmarks in action recognition, UCF101 [8] and HMDB [14] using three ConvNet frameworks (details in Section III.B).

UCF101 is a publicly available dataset, containing 13320 video clips, which is organized in three splits of training and testing data. Our tabulated results contain the average obtained accuracies on these three splits (Table I). HMDB is also an action recognition benchmark dataset containing 68K video clips, which is also organized in splits of training and testing data [14]. The number of outputs for the final fully connected layer in all frameworks equals to the action classes which is 101 and 51 for UCF101 and HMDB datasets, respectively.

### B. Exp. I: Non-biased Background Dataset[1]

The aim of this experiment is to investigate how the state-of-the-art methods [23], [24] and our method perform in situations where the action's background differs from the conventional background that exists in the public dataset. We gathered almost 20 video samples for each of 11 actions, mentioned in Figure 3a, from the real videos recorded by a camera on the QUT campus and some available Youtube video samples in order to include a wider range of context in background compared to the UCF101 dataset (Figure 3a). We tested both ConvNet models [23], [24], trained on UCF101 dataset (provided in Table I), on the new dataset that we named "Non-biased background dataset".

### C. Exp. II: Moving Camera Dataset[2]

In this experiment, we recorded several unconstrained videos using a mobile robot (Guiabot) moving around our work environment to capture students doing normal and abnormal actions in the office environment (Figure 4). This datasets contains 16 videos for each action recorded in four places, office, corridor, kitchen and classroom. Camera motion ranges involved the robot moving from side to side, approaching the subject and rotating around the subject.

### D. Exp. III: Abnormal Behavior Detection

The aim of this experiment is detecting abnormal behavior in workspace environment by a mobile robot. Depending on the environment, different action classes are more likely to be observed than others. For instance, in a robotic lab, we do not expect to see people eating, drinking or playing sports. We propose to exploit such knowledge in our abnormal behavior detection system, which leverages the successes of ConvNets for action recognition and place categorization.

To this end, robot initially requires to identify the place as well as the action being performed by the human. Then, by incorporating the learned prior knowledge, robot makes a decision on whether human behavior in that classified environment is normal or not. We divide our explanation of this task into five stages:

1) Scene categorization: In this part, we aim to do a frame based scene categorization. To this end, we use the Places205 network published by Zhou et al. [31], which is the state-of-the-art in scene categorization and follows the VGGNet architecture for training [29]. Their training dataset contains 2.5 million images of 205 semantic categories, with at least 5,000 images per category. We feed our new dataset recorded on the mobile robot (Section C) into the Places205. The output is a probability distribution over the 205 known scene types and select the highest probability as the probability of the given scene $P(S_i)$.

2) Learning the prior knowledge: our system should learn the likelihood of each scene-action pair, which would enable the robot to make a decision about the

---

[1] This dataset will be publically available.
[2] This dataset will be publically available.

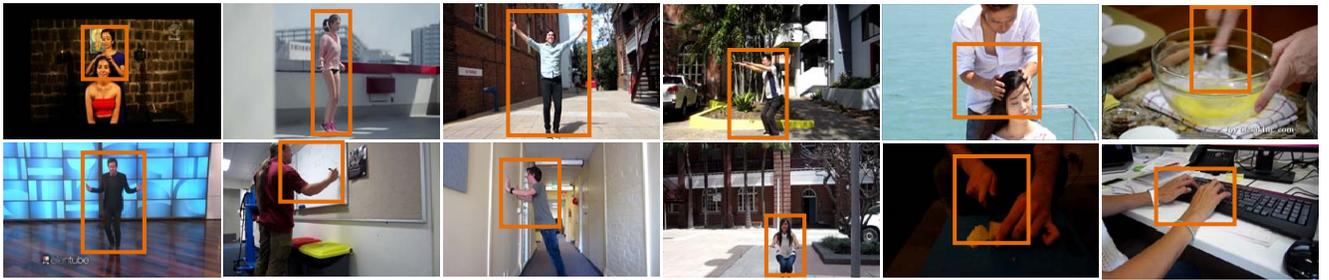

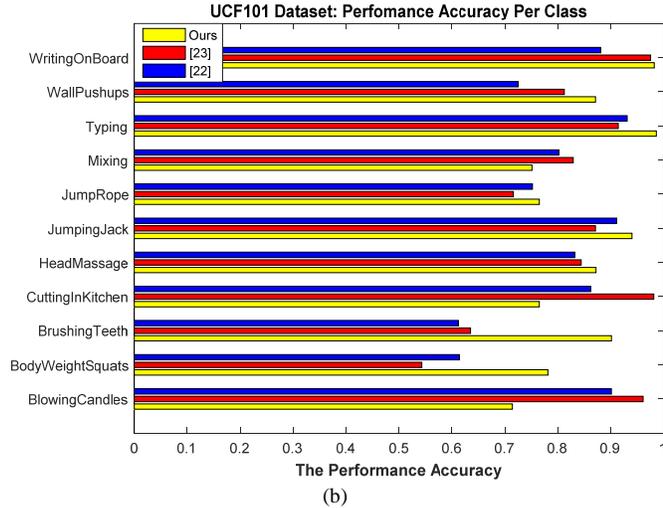
(b)

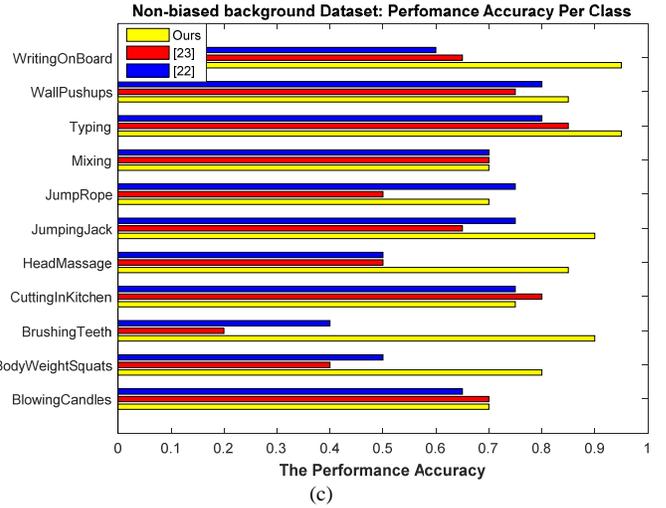
(c)

Figure 3. *(a)* Performance comparison of Three methods on UCF101 dataset. (b) Performance comparison of three methods on Non-biased background dataset. (c) Samples of generated action region proposals on our non-biased background dataset.

normality or abnormality of the human behavior. To this end, we need to calculate occurring likelihood of each action in different scenes in our scenario. We denote this probability as $P(S_i/A_i)$. To compute $P(S_i/A_i)$, we input samples of two public datasets on action recognition, UCF101 and HMDB to the Places205 network and take the scene with maximum probability as the most frequently occurred place for each action.

3) Action recognition regardless of the scene: we denote the probability of the recognized action through our action recognition pipeline as $P(A_i)$.

4) Computing the occurrence probability of actions given the scene: we calculate this likelihood, $P(A_i/S_i)$, for each action and scene using the following equation:

$$P(A_i | S_i) = \frac{P(S_i | A_i) \cdot P(A_i)}{P(S_i)} \quad (2)$$

Where $P(S_i)$, $P(S_i/A_i)$ and $P(A_i)$ can be gained from the first, second and third stages, respectively.

5) Decision making: the aim of this stage is to compare the occurrence probability of an action given a scene $P(A/S)$ (obtained from stage 4) with the occurrence likelihood of the same action with no scene knowledge $P(A)$ (obtained from the stage 3). We follow a simple comparison algorithm; the recognized action in the detected scene is an abnormal behavior if the Abnormal Behavior Detection index, defined as $ABD\_Ind$ in equation (3), returns a positive number greater than a pre-defined threshold, Otherwise, it would be considered as a normal activity.

$$ABD\_Ind = P(A_i) - P(A_i | S_i) \quad (3)$$

Since the problem is a binary classification and the probability values are scattered between [0,1], we set the threshold to 0.5. For instance, if $P(A/S)$ is very low, only a recognized action with probability greater than 0.5 can meet the condition for being an abnormal behavior.

Figure 5 demonstrates the overview of our abnormal behavior detection system and how it performs on one correctly identified example from our dataset.

## V. RESULTS

This section present the results obtained from the experiments described above.

### A. Validation on Benchmarks

In this section, we present the results of our action recognition system on UCF101 and HMDB. Table I provides an extensive comparison with the state-of-the-art methods. We believe the main reason to achieve the matching performance with the state-of-the-art without exploiting the background cues is the elimination of camera motion. We can systematically crop the salient motion areas which leads to a more precise feature learning process.

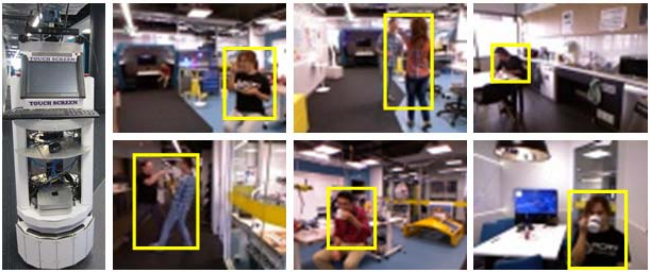

Figure 4. Samples of generated action region proposals and recognized scene and action label on our Guiabot robot dataset.

TABLE I. PERFORMANCE COMPARISON WITH THE STATE-OF-THE-ART DEEP NETWORKS ON UCF101 AND HMDB DATASET

| Methods | UCF101 | | | HMDB |
|---|---|---|---|---|
| | Spatial | Temporal | Full | Spatial |
| Ours with Two-stream Net. | 70.1% | **80.7%** | 88.63% | 40% |
| Ours with C3D | - | - | 73.3% | 40.8% |
| Ours with ResNet | **74.73%** | - | - | 42.1% |
| [22] | 72.7% | 73.9% (L=1) 81% (L=10) | - (L=1) 88% (L=10) | 40.5% |
| [20] | 71.1% | 76.9% | 82.9% | - |
| [21] | 73.1% | - | 88.6% | - |
| [4] | - | - | 65.4% | - |
| C3D on full img. | - | - | 79.8% | 49.91% |
| ResNet on full img. | **79.82%** | - | - | 49.9% |

### B. Exp. I: Non-biased Background Dataset

Figure 3c verifies the outperformance of our method compared to the existing state-of-the-art methods [22], [23], [24], when background does not include any informative context (Non-biased background dataset). Figures 3b and 3c demonstrate the consistency in performance of our method regardless of the background context on both datasets. It is important to note changing the background in our new dataset, negatively impacts the performance of the state-of-the-art methods.

Due to random image cropping in [22], [23] versus selecting the motion salient areas in our approach during the training process, it is more likely that these methods fail to contain the motion cues.

### C. Exp. II: Moving Camera Dataset

This experiment shows how our action recognition system successfully handles the camera motion better than the state-of-the-art methods. Table II demonstrates the accuracies for the proposed models in [22], [23], [24] and our method on our robot dataset using a moving camera. The reason would be due to eliminating the camera motion effects by actively training on action regions rather than full images.

TABLE II. PERFORMANCE COMPARISON OF OUR APPROACH AGAINST [22,23,24] ON GUIABOT ROBOT DATASET

| Actions | Action recognition methods | | | |
|---|---|---|---|---|
| | *Ours* | *[21]* | *[22]* | *[23]* |
| BodyWeightSquats | **100%** | 62.5% | 50% | 75% |
| JumpRope | **93.75%** | 50% | 50% | 50% |
| Punch | **93.75%** | 43.75% | 37.5% | 43.75% |
| Eat | **68.75%** | 6.25% | 18.75% | 18.75% |
| Drink | **81.25%** | 25% | 25% | 31.25% |

### D. Exp. III: Abnormal behavior Detection

The results in this experiment show the power of our proposed approach in Section IV.D in detecting abnormal human behaviors in the workspace.

The system used in this experiment includes an action recognition pipeline, a scene categorization method in addition to learning the prior knowledge. We investigate the use of three state-of-the-art action recognition approaches in the abnormal detection pipeline, while the rest of the system remains the same.

The results indicate an **87.50%** success rate for abnormal human behavior detection on our moving camera dataset containing 16 videos for each action in four places. We test [22], [23], [24] on our dataset. Results are shown in Table III.

TABLE III. COMPARISON OF ABNORMAL BEHAVIOR DETECTION SUCCESS RATES

| *Ours* | *[22]* | *[23]* | *[24]* |
|---|---|---|---|
| **87.50%** | 37.5% | 37.5% | 43.75% |

We conjecture the ability of action recognition method regardless of environment and camera motion plays a significant role in enabling the robot to achieve a higher success rate in detecting abnormal behavior.

## VI. CONCLUSION

In this paper, we focused on two of the main challenges in transitioning from computer vision methods to robotics applications; the sensitivity of many traditional approaches on background cues, and the effect of camera motion in a robotics context.

We addressed these challenges by developing methods for selecting action region proposals that are motion salient and more likely to contain the actions, regardless of background and camera motion. Using two new datasets, the "Non-biased background dataset" and the "Moving camera dataset", we demonstrated our method using both spatial and temporal images to outperform state-of-the-art ConvNet models, and enabling the development of an abnormal behavior detection system. The results obtained indicate how combining a robust action recognition system with the semantic scene category knowledge can enable a robot to detect normal and abnormal human behavior in a typical office environment.

In future work, robots equipped with SLAM systems that have access to semantic information will enable better action recognition performance. Real world robot operation introduces a number of challenges including varying lighting and motion blur; we will adapt successful investigations into learning features that are invariant to these issues in other fields such as place recognition to apply to action recognition. Finally, we plan to investigate the utility of online action recognition for informing robot operations in a range of tasks such as domestic chores and assistive robotics.

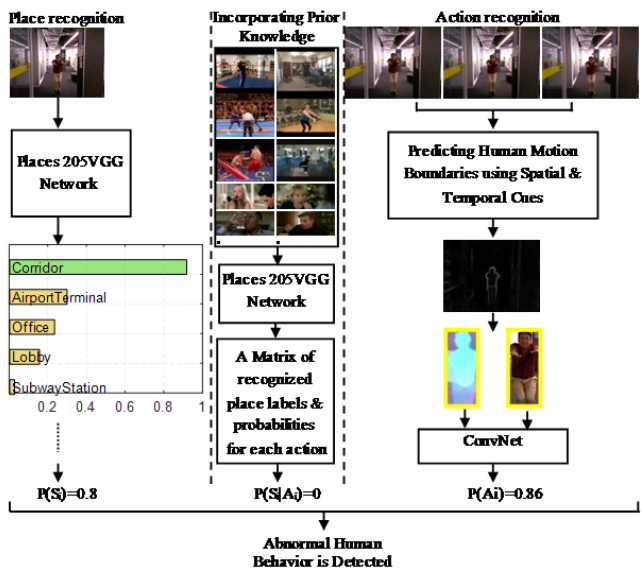

Figure 5. Overview of our approach for unbiased human action recognition on a sample of the Guiabot robot dataset.

ACKNOWLEDGMENT

This Research has supported by a QUTPRA and Australian Centre of Excellence for Robotic Vision (project number CE140100016). I would like to thank Professor Gordon Wyeth who provided insights and expertise that greatly assisted this research.